# Active Learning for Domain Classification in a Commercial Spoken Personal Assistant

*Xi C. Chen, Adithya Sagar, Justine T. Kao, Tony Y. Li,*
*Christopher Klein, Stephen Pulman, Ashish Garg, Jason D. Williams*

Apple Inc.
One Apple Park Way
Cupertino, CA 95014 United States

{xchen6,agurram,jkao,tony_y_li,christopher_klein,spulman,ashishgarg,jason_williams4}@apple.com

## Abstract

We describe a method for selecting relevant new training data for the LSTM-based domain selection component of our personal assistant system. Adding more annotated training data for any ML system typically improves accuracy, but only if it provides examples not already adequately covered in the existing data. However, obtaining, selecting, and labeling relevant data is expensive. This work presents a simple technique that automatically identifies new helpful examples suitable for human annotation. Our experimental results show that the proposed method, compared with random-selection and entropy-based methods, leads to higher accuracy improvements given a fixed annotation budget. Although developed and tested in the setting of a commercial intelligent assistant, the technique is of wider applicability.

## 1. Introduction

In Figure 1 we show the workflow of Siri, a typical speech or text-driven intelligent personal assistant.

Speech or text input is processed so as to recover the intended domain of application (phone call, setting an alarm, querying a calendar, etc.) and then to identify the user's intent, along with any argument slots needed to fulfil the request.

The bottom row of Figure 1 shows the sub-components of the Siri Natural Language Understanding system. To formulate an intent from each request, we first assign the request to a domain using a classifier we call the "Domain Chooser" (DC). (Table 1 shows some example utterances assigned to their most likely domains: there are more than 60 such domains). Once a domain is assigned, we then use a Statistical Parser (SP) to assign a parse label to each token of the utterance. Finally, a post-processing step maps the domain and parse labels predicted by DC and SP into an intent representation and sends it to the Actions component to perform the appropriate action.

We focus here on the "Domain Chooser" component, a multi-class system which uses an ensemble of Bidirectional Long Short Term Memory (BiLSTM) units. All members of the ensemble share the same architecture (as shown later in Figure 3) but use varying hyper-parameters to optimize various specific metrics. The resultant probability of the ensemble is calculated as the geometric mean of the probabilities from individual models.

As with any machine learning system, the quality of training data for the DC component is of vital importance. In our setting, with new domains frequently being added, and the cov-erage of existing domains often shifting over time with current events or trends in popular culture, we have a constant need for new training examples, either to improve accuracy on existing coverage or to extend to new phenomena. However, we need to make sure that we add only relevant and helpful data, for several reasons: the data needs to be human-annotated, which is a slow and expensive process, and we need to be sure that the data is going to correct or extend the classifiers rather than just reinforcing what they have already learned. Active learning [1] methods are widely used in these circumstances, as ways of identifing relevant new training data examples suitable for human annotation. The method we propose can be seen as another type of active learning technique.

Table 1: *Example utterances with their domain labels*

| Utterance | Domain |
|---|---|
| What is the weather today | Weather |
| Make a payment to Xi Chen | Payment |
| Call Chris | Phone |
| Play Hello from Adele | Music |
| What time does Starbucks close | LocalBusiness |
| What time is Thanksgiving day parade | LocalEvent |

## 2. Related work

[1, 2] provides a good overview of active learning methods. In the context of natural language processing, perhaps the most widely used in practice are random sampling (which can improve things, but is demonstrably inferior to more motivated sampling schemes); "human-in-the-loop" techniques, in which an expert is presented with examples wrongly labelled by the existing system and decides which of them should be correctly annotated and added to the training data (for a recent example see [3]); various types of "uncertainty sampling" in which probabilistic measures like entropy or "least confidence" are used to identify candidates to be annotated, e.g. [4, 5, 6]; "query by committee", in which an ensemble of classifiers label the same example and cases where there is maximum disagreement are taken to be the relevant candidates; or, more recently, approaches using reinforcement learning to reward "good" choices of new candidates: [7, 8, 9]. Work with similar aims and approach to ours are reported in [10], which appeared after this paper was submitted. It would be interesting to compare these two methods against each other, but regrettably, in both cases the data is for privacy reasons not publicly available.

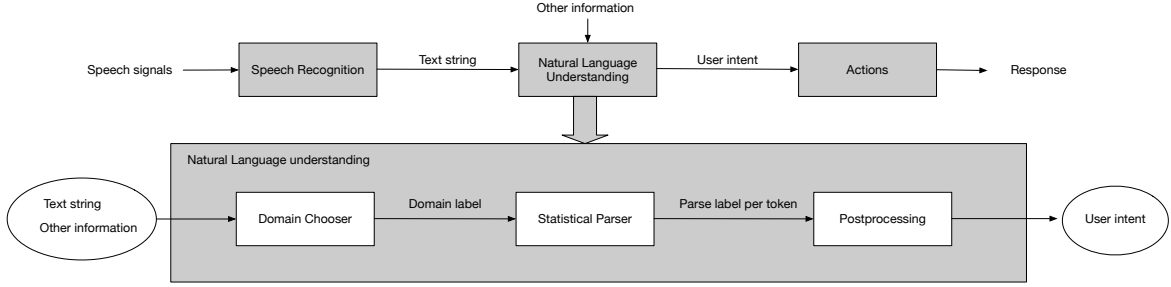

Figure 1: *Simplified work flow of Siri.*

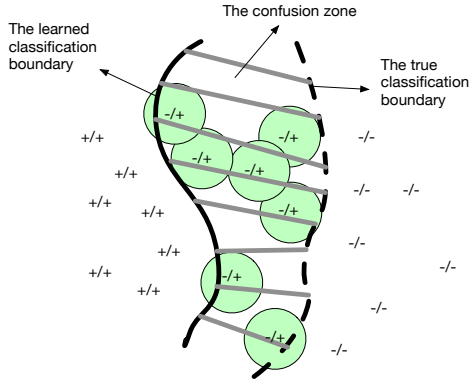

Figure 2: *The solid line is the learned classification boundary, the dotted line is the true classification boundary, and the area in between is the confusion zone. Data located in the confusion zone are marked as -/+ since their predicted label is - and their ground truth label is +. The green circles indicate the area where candidate training data is located.*

## 3. The proposed method

In our setting, human-in-the-loop or reinforcement learning approaches are not practical. Uncertainty sampling is feasible, as is query-by-committee, and we propose a method which has some features of both. As with uncertainty sampling, we want to identify those incorrectly labeled candidates that are near to the decision boundary learned by the existing classifier, so that including correctly labeled versions will sharpen the boundaries learned on the new data. Like query-by-committee, we use an ensemble of classifiers, but rather than using the ensemble to identify a single candidate, we use their results to suggest distinct new candidates for each classifier, which are then retrained separately.

In more detail, our proposed method has three steps. Firstly, we want to estimate the decision or classification boundary learned by the existing classifier. The candidate data points of interest to us are those that are located in what we call the "confusion zone", namely the area between the learned classification boundary, and the true one. We can picture this abstractly as in 2.

In the confusion zone, the predicted labels do not match the ground truth labels. Prediction errors therefore suggest the location of the confusion zone, and since the confusion zone is an estimation of the classification or decision boundary, prediction errors can be used to estimate the decision boundaries of the classifier.

A model deployed in a commercial production system allows for relatively straightforward discovery of prediction errors. Common ways to get the prediction errors are from bug reports, from methodical testing by quality assurance teams, and from users' negative engagement signals, i.e. actions by the user that imply that their request was not correctly processed, such as early termination of a piece of music, or switch to a direct use of an application.

However, although these prediction errors are themselves very valuable, they are relatively few in number. The second step of our method is to expand the set of hypothesised prediction errors by finding examples in a pool of unlabeled data that are similar to the confirmed prediction errors. To do this, we use a k-Nearest Neighbor (kNN) method, since that makes it easy to control the total number of selected training data examples.

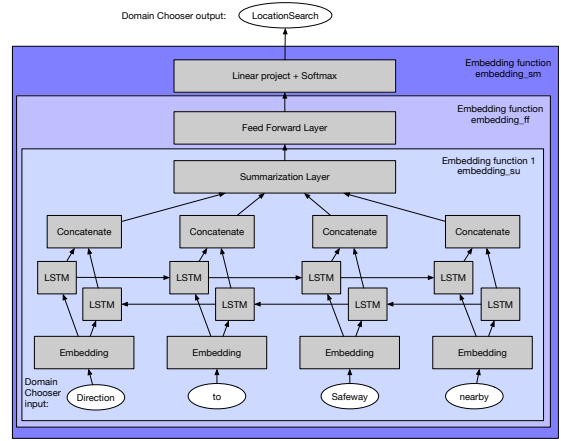

Figure 3: *The embedding function we used to query for candidate training data.*

As, mentioned, the input data to our DC model is a combination of text string, named entity and contextual information. To find data that are near the classification boundary, we need to define a similarity measure and representations of the data that can be input to that measure. The similarity function we use is simple Euclidean distance over vector representations of DC inputs, derived from the embeddings produced by the BiLSTMs themselves. Figure 3 shows the embedding functions we use. In detail, we construct three embedding functions: the input to the summarization layer $embedding_{su}$, the input to the feed forward layers $embedding_{ff}$, and the input to the softmax layer $embedding_{sm}$. A data point $d$ is considered as a kNN neighbor of a prediction error $e$ if $d$ is the $k$-nearest-neighbor of $e$ under any embedding function.

Once we have found a set of potential candidates, the third step is simply to get human annotators to assign the correct labels and add the result to the original training data.

Our DC model is an ensemble of 7 individual BiLSTM models. Since each classifier is equally important to our system, we aim to improve the accuracy of each individual model using the proposed method, independently. For example, if an utterance yields an error from one member model, we will usually add up to $3 \times k$ new training examples (k training examples from each of the three layers) which are nearest to it in the unlabeled pool. If instead 5 member models make an error, we would add up to $5 \times 3 \times k$ training data items from that utterance, since the embeddings for each model may well pick out different candidates.

### 3.1. Prediction errors and their kNNs

Clearly, discovery of similar utterances to the prediction errors is a key component in the proposed method. As a sanity check, we manually inspect a sample for the similarity between the prediction errors and their $k$ nearest neighbors. We find that there are generally clear connections between the prediction errors and their corresponding suggested training examples (see the samples below). In many cases, the ground truth labels of selected training data from the same prediction are different. This fact suggests that the candidate training data are indeed largely being sampled from the confusion zone.

**Prediction error 1 :**

- Charge my phone to hundred percent (*Settings*)

**Similar examples found :**

- Low Power Mode (*Settings*)
- I am I need you to charge my phone (*Settings*)
- Turn my phone off (*Unsupported function*)
- Can you switch off my phone (*Unsupported function*)
- Pair my Bluetooth (*Settings*)

**Prediction error 2 :**

- Spell volume (*Settings*)

**Similar examples found :**

- Increase volume (*Settings*)
- What is the volume (*Settings*)
- Spell squawk (*Dictionary*)
- Spell climate (*Dictionary*)

**Prediction error 3 :**

- What day is Brooklyn Nine-Nine on (*TV*)

**Similar examples found :**

- What day is Easter on (*Clock*)
- What day is the third on (*Clock*)
- What is the week is January 1 fall on (*Clock*)
- When is election day (*Politics*)

## 4. Experiments

To assess the effectiveness of the proposed method, we evaluate it in several ways. Firstly, we assess its ability to enhance the accuracy of an in-production virtual assistant system as compared to existing training data: does this method produce better quality examples?

Secondly, we compare it with an entropy-based uncertainty sampling method to see which results in a better performance improvement. In addition, we also evaluate the cost of the two methods due to the existence of "out of domain" examples. "Out of domain" examples are those which our current system cannot deal with even in principle, for example "Book me a hotel this weekend on the planet Mars", or which convey an unclear user intent, probably as the result of ASR errors, such as "Can you forget that private with 10%". Out-of-domain utterances are common in commercial personal assistants, but because they are highly varied, it is difficult for a multi-class classifier alone to identify them. Instead, specialized techniques using a separate classifier or signals from other downstream components are often used. In our setting, our targeted classifier does not have a special "out-of-domain" class, and so obtaining labels of the "out-of-domain" utterances is a waste of grading resources, since they cannot help improve the target classifier. We are therefore interested to know what fraction of utterances retrieved by different methods are "out-of-domain".

### 4.1. Experiment I: baseline training set vs augmented training set

We use a blind test set to evaluate the performance of different methods. The prediction errors are derived from a development set, which has been used previously to debug the system. In this experiment, we demonstrate that training data obtained by the proposed method is better than the existing training data. We call the existing training data the baseline training set. The baseline training set we used is a combination of data that are randomly sampled from usage logs and data that have been manually generated by engineers in developing the features of the assistant system.

The size of the baseline training data we used is $870k$. To compare the quality of training data obtained using the proposed method, we randomly selected $850k$ utterances from the baseline training set and we add $20k$ labeled training data that was obtained using our method. We train the same model architecture (as shown in Figure 3) with the two different training data sets. We run the experiments 11 times and take the mean error rate for each data set. Figure 4 shows the performance of the models that are trained by the baseline data (left) and the augmented data (right). We find that, in this scenario, swapping only $2.3\%$ of the original training data reduces the model's error rate by $8.88\%$.

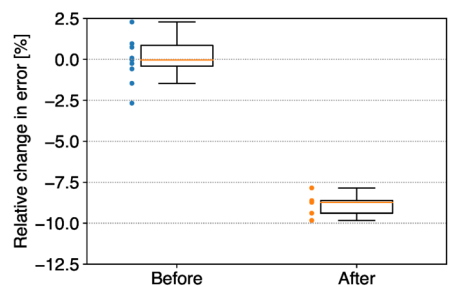

Figure 4: *Blind test error rates with the baseline training data (left) and the newly augmented training data (right). The relative error reduction is 8.88% (p = 3.69e-10)*

## 4.2. Experiment II: entropy-based method vs the proposed method

Uncertainty sampling using entropy or some similar method is often considered as the state-of-art choice for training data selection. In this experiment we compare our method with an entropy-based sampler. The entropy-based method we use here selects usage data with high entropy scores (suggesting model uncertainty) as candidate training data. The entropy score is calculated using the probability distribution over the set of labels that is provided by the softmax output layer of the model.

We created a training data set of 10k new utterances by this method. Regrettably, resource and time constraints meant that a larger entropy-based data set, comparable in size to the similarity-based one used earlier, was not available.

Next, we created an entropy-augmented-training-set and a similarity-augmented-training-set by adding the new utterances into the $850k$ baseline training set. The size of the two training sets we compared here are therefore $860k$. In other words, we add $1.18\%$ of data using two methods. As in Experiment I, we train the same model architecture with different training data 11 times and take the mean of the error rates for each data set. The results (Figure 5) shows that we get around $3.11\%$ relative error reduction by using the entropy-based method and get around $3.38\%$ relative error rate reduction by using our proposed similarity-based method. This difference is not as dramatic as the earlier comparison, probably because the proportion of new data is so small.

However, along another dimension of comparison, own method is considerably superior to the entropy based method.

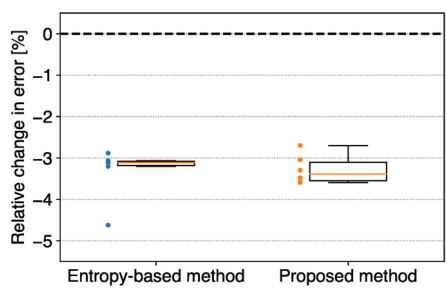

Figure 5: *Blind test error rate reduction with training data obtained using the entropy based method (left) and the training data obtained using our proposed method (right).*

In order to obtain 10k usable (i.e. not out of domain) examples, our annotators needed to label 12.8k utterances obtained by our method, whereas with the entropy based method, our annotators needed to label as many as 18.7k utterances. Figure 6 shows the out-of-domain rate of the proposed method and the entropy based method. Our method is much more efficient at finding usable examples: an important consideration given how slow and costly it is for humans to annotate such training data.

## 5. Properties of the derived training data

We noticed some interesting properties of the relationship between the new training data harvested by our method and the original prediction error used to get this data. Firstly, as was seen in 3.1, examples that are similar to the prediction error may correctly be assigned several different domain labels when inspected by human annotators. What happens to the original prediction error under these circumstances? Perhaps not surprisingly, we found that the original error was more likely to be corrected when training on the new data if the nearby examples found shared the same label with it.

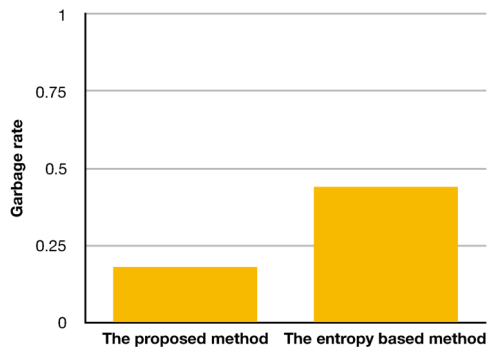

Figure 6: *The out-of-domain rate of the training data selected from the entropy based method and our proposed method. The result shows that using the entropy based method causes a huge waste of annotation resources since $44\%$ of the suggested candidates are out-of-domain.*

Secondly, the relation between the number of new training examples found and success in correcting the original prediction error is more complex than we might have expected. Pre-theoretically, one might think that the more data, the more likely the error is to be corrected. The number of new training examples found for a particular prediction error can vary: if each of the three layers of a model used suggests different candidates, and if each of seven models does likewise then we might get up to k×21 new examples. In practice we found that more data was better, but only up to a certain point: if a large number (80+) examples was used, the prediction error was unlikely to be corrected. We hypothesise that several factors may explain this: if more models were predicting the example incorrectly and inconsistently, it is clearly a difficult example to classify; and if different layers of the same model are finding different examples, that suggests that the hidden states involved display a lot of variation, again suggesting uncertainty and inherent difficulty.

## 6. Conclusions

In this paper, we have proposed a simple but effective method for efficient discovery of useful training data for a domain chooser classifier, as part of a commercial spoken personal assistant system. After applying the proposed method to our existing system, we were able to demonstrate significant improvement in an independently generated blind test set. In particular, we have shown that on the blind test set, we can achieve an $4.51\%$ reduction on errors by adding only $2.3\%$ new training data examples. The method produces also produces better quality data than an uncertainty sampling technique, as shown by the increased accuracy of the classifier, and the lower rate of "useless" out of domain examples suggested. This reduces the time taken for human annotation.